\documentclass{article}
\usepackage{spconf,amsmath,epsfig}
\usepackage{booktabs}
\usepackage{graphicx}
\usepackage{threeparttable}
\usepackage{bbm}
\usepackage{amssymb}
\usepackage[ruled,vlined,linesnumbered]{algorithm2e}
\usepackage{setspace}  
\usepackage{textcomp}
\usepackage{caption}

\begin{document}\sloppy

\def\x{{\mathbf x}}
\def\L{{\cal L}}

\title{Real-time Multiple People Tracking with Deeply Learned Candidate Selection and Person Re-Identification}
%
\name{Long Chen, Haizhou Ai, Zijie Zhuang, Chong Shang}
\address{
Tsinghua National Lab for Info. Sci. \& Tech. (TNList), \\
Department of Computer Science and Technology, Tsinghua University, Beijing, China, 100084. \\
l-chen16@mails.tsinghua.edu.cn
}

\maketitle

\begin{abstract}
Online multi-object tracking is a fundamental problem
in time-critical video analysis applications.
A major challenge in the popular tracking-by-detection framework is 
how to associate unreliable detection results with existing tracks.
In this paper, 
we propose to handle unreliable detection by
collecting candidates from outputs of both detection and tracking.
The intuition behind generating redundant candidates
is that detection and tracks can complement each other in different scenarios.
Detection results of high confidence prevent tracking drifts in the long term,
and predictions of tracks can handle noisy detection caused by occlusion.
In order to apply optimal selection from a considerable amount of candidates in real-time,
we present a novel scoring function based on a fully convolutional neural network,
that shares most computations on the entire image.
Moreover, we adopt a deeply learned appearance representation, 
which is trained on large-scale person re-identification datasets,
to improve the identification ability of our tracker.
Extensive experiments 
show that our tracker achieves real-time and state-of-the-art performance
on a widely used people tracking benchmark.

\end{abstract}
\section{Introduction}
\label{sec:intro}

Tracking multiple objects in a complex scene is a
challenging problem in many video analysis and multimedia applications, such as
visual surveillance, sport analysis, and autonomous driving.
The objective of multi-object tracking is to estimate trajectories of objects in a specific category.
Here we tackle the problem of people tracking by taking advantage of person re-identification.

\begin{figure}[t]
\begin{center}
\includegraphics[width=0.95\linewidth]{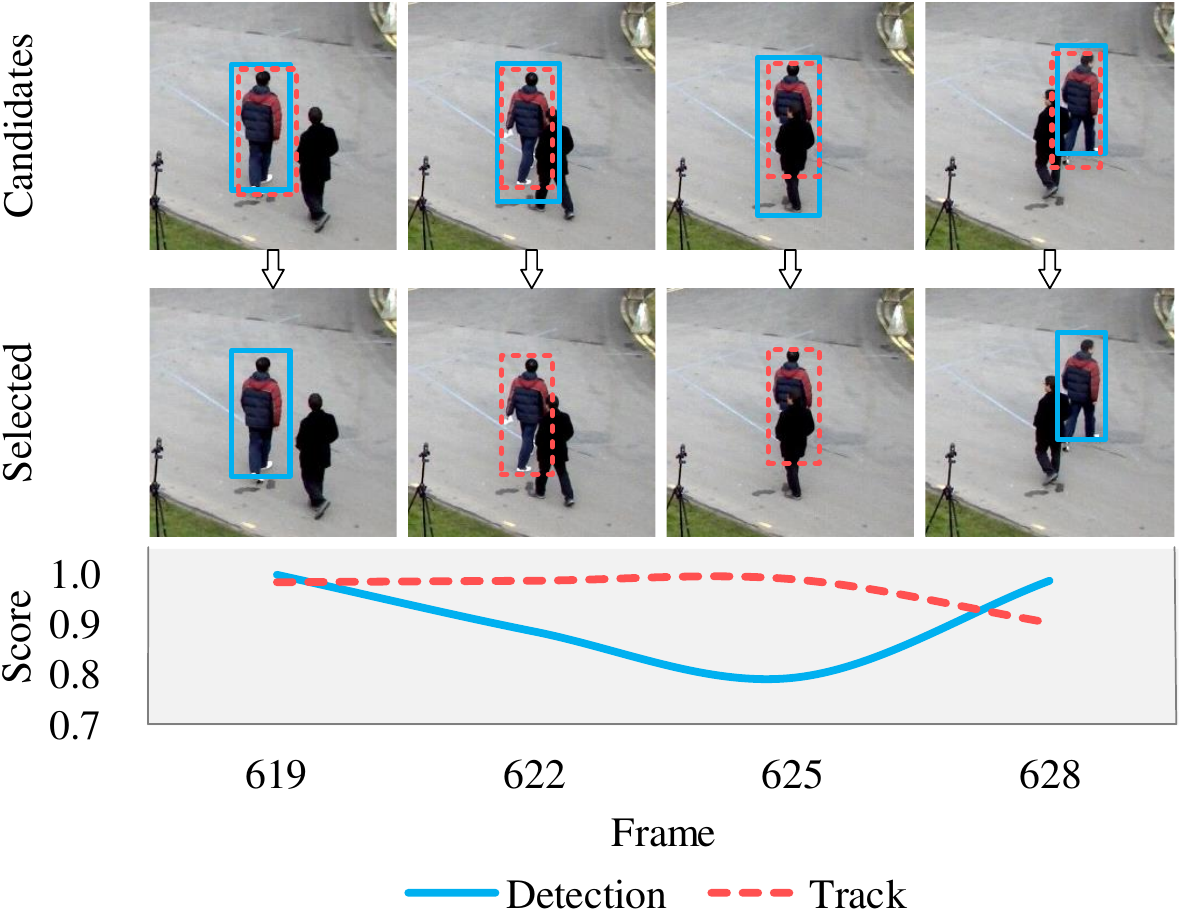}
\end{center}
\caption{
Candidate selection based on unified scores.
Candidates from detection and tracks are visualized as blue solid rectangles and red dotted rectangles, respectively.
Detection and tracks can complement each other for data association. }
\label{fig:det-trk}
\end{figure}

Multi-object tracking benefits a lot from advances in object detection in the past decade.
The popular tracking-by-detection methods apply the detector on each frame, and associate 
detection across frames to generate object trajectories.
Both intra-category occlusion and unreliable detection are tremendous challenges in such a tracking framework \cite{bae2017confidence,fagot2016improving}.
Intra-category occlusion and similar appearances of objects can result in ambiguities in data association.
Multiple cues, including motion, shape and object appearances, are fused to
mitigate this problem \cite{sadeghian2017tracking,yan2012track}.
On the other hand, detection results are not always reliable.
Pose variation and occlusion
in crowded scenes often cause
detection failures such as false positives, missing detection, and non-accurate bounding.
Some studies proposed to handle unreliable detection in a batch mode \cite{fagot2016improving,kim2015multiple,tang2017multiple}.
These methods address detection noises by introducing information from future frames.
Detection results in whole video frames or a temporal window are employed and linked to trajectories 
by solving a global optimization problem.
Tracking in a batch mode is non-causal and not suitable for time-critical applications.
In contrast to these works, we focus on the online multiple people tracking problem, using only the current and past frames.



In order to handle unreliable detection in an online mode,
our tracking framework optimally selects candidates 
from outputs of both detection and tracks in each frame (as shown in Figure \ref{fig:det-trk}).
In most of the existing tracking-by-detection methods, when talking about data association,
candidates to be associated with existing tracks are only made up of detection results. 
Yan et al. \cite{yan2012track} proposed to treat the tracker and object detector as two independent identities,
and keep results of them as candidates.
They selected candidates based on hand-crafted features, 
e.g., color histogram, optical flow, and motion features.
The intuition behind generating redundant candidates
is that detection and tracks can complement each other in different scenarios.
On the one hand,
reliable predictions from the tracker can be used for short-term association in case of  
missing detection or non-accurate bounding.
On the other hand,
confident detection results are essential to prevent tracks drifting to backgrounds in the long term.
How to score outputs of both detection and tracks 
in an unified way is still an open question.

Recently, deep neural networks, especially convolutional neural networks (CNN), 
have made great progress in the field of computer vision and multimedia.
In this paper, we take full advantage of deep neural networks 
to tackle unreliable detection and intra-category occlusion.
Our contribution is three fold.
First, we handle unreliable detection in online tracking by 
combining both detection and tracking results as candidates,
and selecting optimal candidates based on deep neural networks.
Second, we present a hierarchical data association strategy, which utilizes spatial information and deeply learned person re-identification (ReID) features.
Third, we demonstrate real-time and state-of-the-art performance of our tracker on a widely used people tracking benchmark.


\section{Related Work}
Tracking-by-detection is becoming the most popular strategy for multi-object tracking.
Bae et al. \cite{bae2017confidence} associated tracklets with detection in different ways
according to their confidence values.
Sanchez-Matilla et al. \cite{sanchez2016online} exploited multiple detectors 
to improve tracking performance.
They collected outputs from multiple detectors, during the so called over-detection process.
Combining results from multiple detectors can improve the tracking performance 
but is not efficient for real-time applications.
In contrast, our tracking framework needs only one detector and
generates candidates from existing tracks.
Chu et al. \cite{chu2017online} used a binary classifier and single object tracker for online multi-object tracking.
They shared the feature maps for classification but still had a high computation complicity.

Batch methods formulate tracking as a global optimization problem \cite{yan2012track,kim2015multiple,tang2017multiple,guan2014multi}.
These methods utilized information from future frames
to handle noisy detection and reduce ambiguities in data association.
Liu et al. \cite{liu2017rewind} proposed a rewind to track strategy to generate backward tracklets involving future information,
to obtain a more stable similarity measurement for association.
Person re-identification was also explored in \cite{tang2017multiple,guan2014multi,tang2016multi} for the global optimization.
Our framework leverages deeply learned ReID features in an online mode, 
to improve the identification ability when coping with the problem of intra-category occlusion.

\section{Proposed Method}

\subsection{Framework Overview}
In this work,
we extend traditional tracking-by-detection
by collecting candidates from outputs of both detection and tracks.
Our framework consists of two sequential tasks, that is,
candidate selection and data association.

We first measure all the candidates using an unified scoring function.
A discriminatively trained object classifier 
and a well-designed tracklet confidence 
are fused to formulate the scoring function,
as described in Section \ref{sec:rfcn} and Section \ref{sec:conf}.
Non-maximal suppression (NMS) is subsequently performed with the estimated scores.
After obtaining candidates without redundancy,
we use both appearance representations and spatial information 
to hierarchically associate existing tracks with the selected candidates.
%
%
Our appearance representations are deeply learned from the person re-identification as described in Section \ref{sec:reid}.
Hierarchical data association is detailed in Section \ref{sec:associate}.


\subsection{Real-Time Object Classification}
\label{sec:rfcn}
Combining outputs of both detection and tracks will result in an excessive amount of candidates.
Our classifier shares most computations on the entire image 
by using a region-based fully convolutional neural network (R-FCN) \cite{dai2016r}.
Thus, it is much more efficient comparing to classification
on image patches, which are cropped from heavily overlapped candidate regions.
The comparison of time consumption of these two methods can be found in Figure \ref{fig:time}.

Our efficient classifier is illustrated in Figure 2.
Given an image frame,
score maps of the entire image are predicted 
using a fully convolutional neural network with an encoder-decoder architecture.
The encoder part is a light-weight convolutional backbone for real-time performance,
and we introduce the decoder part with up-sampling to 
increase the spatial resolution of output score maps for later classification.
Each candidate to be classified is defined as a region of interest (RoI) by ${\bf{x}} = (x_0, y_0, w, h)$,
where $(x_0, y_0)$ denotes the top-left point and $w$, $h$ represent width and height of the region.
For computational efficiency,
we expect that
the classification probability of each RoI is directly voted by the shared score maps.
A straightforward approach for voting is to construct foreground probabilities for all points on the image,
and then calculate the average probability of points inside the RoI.
However, this simple strategy loses the spatial information of objects.
For instance, even if the RoI only covers a part of the object,
a high confidence score still can be obtained.

In order to explicitly encode spatial information into the score maps, 
we employ the position-sensitive RoI pooling layer \cite{dai2016r}
and estimate the classification probability from $k^2$ position-sensitive score maps ${\mathbf{z}}$.
In particular, we split a RoI into $k\times k$ bins by a regular grid.
Each of the bins has the same size $\left \lfloor \frac{w}{k} \times \frac{h}{k} \right \rfloor$,
and represents a specific spatial location of the object.
We extract responses of $k \times k$ bins from $k^2$ score maps.
Each score map only corresponds to one bin.
The final classification probability of a RoI $\bf{x}$ is formulated as:
\begin{equation} \label{eq:cls}
p(y|{\bf{z},\bf{x}})=\sigma (\frac{1}{wh}\sum_{i=1}^{k^2}\sum_{(x,y)\in bin_i}{\bf{z}}_i(x,y)),
\end{equation}
where $\sigma(x)=\frac{1}{1+e^{-x}}$ is the sigmoid function, and ${\bf{z}}_i$ denotes the $i$-th score map.

During the training procedure, 
we randomly sample RoIs around the ground truth bounding boxes as positive examples,
and take the same number of RoIs from backgrounds as negative examples. 
By training the network end-to-end, 
the output on the top of the decoder part, that is,
the $k^2$ score maps, learns to response to specific spatial locations of the object.
%
For example, if $k=3$, 
we have 9 score maps response to top-left, top-center, top-right, ..., bottom-right of the object, respectively.
In this way, the RoI pooling layer is sensitive to spatial positions 
and has a strong discriminative ability for object classification without using learnable parameters.
Please note that the proposed neural network is trained only for candidate classification, 
not for bounding box regression.

\begin{figure}[t]
\begin{center}
\includegraphics[width=0.95\linewidth]{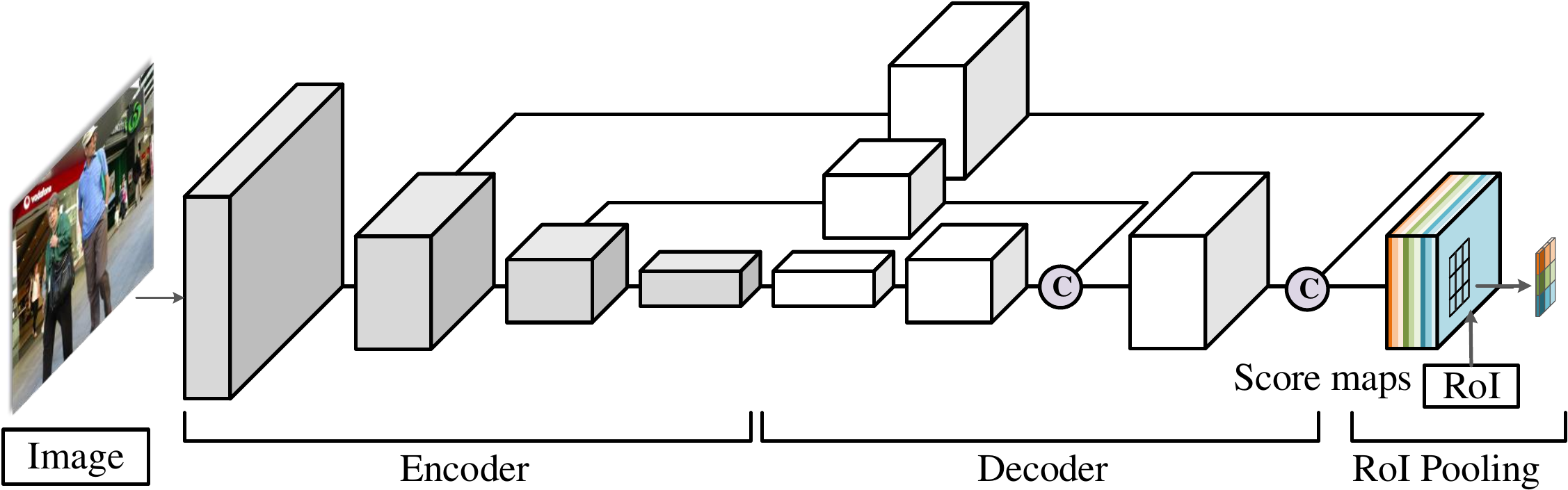}
\end{center}
\caption{
R-FCN architecture for efficient classification.
Features from the encoder network are concatenated with up-sampled features in the decoder part,
to capture both the semantic and low-level information.
Each color in the last block represents a specific score map.
}
\label{fig:network}
\end{figure}

\subsection{Tracklet Confidence and Scoring Function}
\label{sec:conf}
Given a new frame, 
we estimate the new location of each existing track using the Kalman filter.
These predictions
are adopted to handle detection failures caused by varying visual properties of objects and occlusion in crowded scenes.
But they are not suitable for long-term tracking.
The accuracy of the Kalman filter could decrease if it is not updated by detection over a long time.
Tracklet confidence is designed to measure the accuracy of the filter using temporal information.

A tracklet is generated through temporal association
of candidates from consecutive frames.
We can split a track into a set of tracklets 
since a track can be interrupted and retrieved for times during its lifetime.
Every time a track is retrieved from lost state,
the Kalman filter will be reinitialized.
Therefore, only the information of the last tracklet is utilized to formulate the confidence of the track.
Here we define $L_{det}$ as the number of detection results associated to the tracklet, 
and $L_{trk}$ as the number of track predictions after the last detection is associated.
The tracklet confidence is defined as:
\begin{equation} \label{eq:tracklet}
s_{trk}=\mathrm{max}(1- log(1+\alpha \cdot L_{trk}), 0)\cdot \mathbbm{1}(L_{det}\geq 2),
\end{equation}
where $\mathbbm{1}(\cdot)$ is the indicator function that equals 1 if the input is
$true$, otherwise equals 0.
We require $L_{det}\geq 2$ to construct a reasonable motion model using observed detection
before the track is used as a candidate.

The unified scoring function for a candidate $\bf{x}$ is formated 
by fusing classification probability and tracklet confidence:
\begin{equation} \label{eq:score}
s = p(y|{\bf{z},\bf{x}}) \cdot (\mathbbm{1}({\bf{x}}\in C_{det})) + s_{trk} \mathbbm{1}({\bf{x}}\in C_{trk})).
\end{equation}
Here we use ${C}_{det}$ to denote the candidates from detection, and ${C}_{trk}$ for candidates from tracks,
and $s_{trk} \in [0,1]$ to punish candidates from uncertain tracks.
Candidates for data association are finally selected based on the unified scores using non-maximal suppression.
We define the maximum intersection over union (IoU) by a threshold $\tau_{nms}$,
also there is a threshold $\tau_{s}$ for the minimum score.


\subsection{Appearance Representation with ReID Features}
\label{sec:reid}

The similarity function between candidates is the key component of data association. 
We argue that the object appearance, which are deeply learned by a data driven approach,
outperforms traditional hand-crafted features on the task of similarity estimation.
For the purpose of learning the object appearance and similarity function, 
we employ a deep neural network to extract feature vectors from RGB images,
and formulate the similarity using the distance between the obtained features. 

We utilize the network architecture proposed in \cite{zhao2017deeply}
and train the network on a combination of several large scale person re-identification datasets.
The network $H_{reid}$ consists of the convolutional backbone from GoogLeNet \cite{szegedy2015going} followed by $K$ branches of part-aligned fully connected (FC) layers.
We refer to \cite{zhao2017deeply} for more details on the network architecture.
Given an RGB image $\mathbf{I}$ of a person,
the appearance representation is formulated as $f=H_{reid}(\mathbf{I})$.
We directly use Euclidean distance between the feature vectors to measure the distance $d_{ij}$ of two images ${\bf{I}}_i$ and ${\bf{I}}_j$.
During the training procedure, images of identities in training datasets 
are formed as a set of triplets $T=\{\langle {\bf{I}}_i, {\bf{I}}_j, {\bf{I}}_k \rangle \}$,
where $\langle {\bf{I}}_i, {\bf{I}}_j \rangle$ is a positive pair from the same person,
and $\langle {\bf{I}}_i, {\bf{I}}_k \rangle$ is the negative pair from two different people.
Given $N$ triplets, the loss function going to be minimized is formulated as:
\begin{equation} \label{eq:triplet}
l_{triplet}=\frac{1}{N}\sum_{\langle \mathbf{I}_i,\mathbf{I}_j,\mathbf{I}_k \rangle \in\mathbf{T}} \mathrm{max}(d_{ij}-d_{ik} + m, 0),
\end{equation}
where $m > 0$ is a predefined margin.
We ignore triplets that are easy to handle, i.e. $d_{ik} - d_{ij} > m$, 
to enhance the discriminative ability of learned feature representations.

\subsection{Hierarchical Data Association}
\label{sec:associate}
Predictions of tracks are utilized to handle missing detection occurred in crowded scenes.
Influenced by intra-category occlusion,
these predictions may be involved with other objects. 
To avoid taking other unwanted objects and backgrounds into appearance representations,
we hierarchically associate tracks with different candidates using different features.

In particular, we first apply data association on candidates from detection,
using appearance representations with a threshold $\tau_d$ for the maximum distance.
Then, we associate the remaining candidates with unassociated tracks based on 
IoU between candidates and tracks, with a threshold $\tau_{iou}$.
We only update appearance representations of tracks when they are associated to detection.
The updating is conducted by saving ReID features from the associated detection.
Finally, new tracks are initialized based on the remaining detection results.
The detail of the proposed online tracking algorithm
is illustrated in Algorithm 1. 
With the hierarchical data association,
we only need to extract ReID features for candidates from detection once per frame.
Combining this with the previous efficient scoring function and tracklet confidences,
our framework can run at real-time speed.


\section{Experiments}
\label{sec:exp}

\subsection{Experiment Setup}
To evaluate the performance of the proposed online tracking method,
we conduct extensive experiments on the MOT16 dataset \cite{milan2016mot16},
which is a widely used benchmark for multiple people tracking.
This dataset contains a training set and a test set, 
each with 7 challenging video sequences filmed in unconstrained environments.
We form a validation set with 5 video sequences from the training set
to analyze the contribution of each component in our framework.
Afterwards, we submit the tracking result on the test set to the benchmark, and
compare it with state-of-the-art methods on the benchmark.

\textbf{Implementation details.}
We employ SqueezeNet \cite{iandola2016squeezenet}, 
as the backbone of R-FCN for real-time performance.
Our fully convolutional network, consisting of SqueezeNet and the decoder part,
costs only 8ms to estimate score maps for an input image with the size of 1152x640 on a GTX1080Ti GPU.
We set $k=7$ for position-sensitive score maps, and train the network
using RMSprop optimizer with the learning rate of 1e-4 and the batch size of 32 for 20k iterations.
The training data for person classification is collected from MS COCO \cite{lin2014microsoft}
and the remaining two training video sequences.
We set $\tau_{nms}=0.3$ and $\tau_s=0.4$ for candidate selection.
When coping with the ReID network, 
we train it on a combination of three large scale person re-identification datasets,
i.e. Market1501 \cite{zheng2015scalable}, CUHK01 and CUHK03 \cite{li2014deepreid}, to enhance the generation ability for tracking.
We set $\tau_d=0.4$ and $\tau_{iou}=0.3$ for hierarchical data association.
The following experiments are based on the same hyper-parameters.

\textbf{Evaluation metrics.}
In order to measure accuracies of bounding boxes and identities at the same time,
we adopt multiple metrics used in the benchmark to evaluate the proposed method, 
including multiple object tracking accuracy (MOTA) \cite{bernardin2008evaluating}, 
false alarm per frame (FAF), the number of
mostly tracked targets (MT, $> 80\%$ recovered), the number of
mostly lost targets (ML, $< 20\%$ recovered) \cite{li2009learning}, false positives (FP), false negatives (FN),
identity switches (IDS), identification recall (IDR), 
identification F1 score (IDF1) \cite{ristani2016performance}, and processing speed (frames per second, FPS).

\begin{algorithm}[t]
\SetAlgoLined
\DontPrintSemicolon
\SetNoFillComment
\footnotesize
\KwIn{A video sequence $v$ with $N_v$ frames and object detection $\{\mathcal{D}_k\}_{k=1}^{N_v}$}
\KwOut{Tracks $\mathcal{T}$ of the video}

Initialization: $\mathcal{T} \leftarrow \emptyset$; appearance of tracks $\mathcal{F}_{trk} \leftarrow \emptyset$ \;
\ForEach{frame $f_k$ in $v$}{
	Estimate score maps $\bf{z}$ from $f$ using R-FCN \;
	\tcc{collect candidates}
	$C_{det} \leftarrow \mathcal{D}_k$;
	$C_{trk} \leftarrow \emptyset$ \;
	\ForEach{$t$ in $\mathcal{T}$}{
	Predict new location $\mathbf{x}^*$ of $t$ using Kalman filter\;
	$C_{trk} \leftarrow  C_{trk} \cup \{\mathbf{x}^*\}$ \;
	}
	\tcc{select candidates} 
	$C \leftarrow C_{det} \cup C_{trk}$ \;
	$S \leftarrow \text{unified scores computed from Equation \ref{eq:score}}$ \;
	$C, S \leftarrow \texttt{NMS}(C, S, \tau_{nms})$ \;
	$C, S \leftarrow \texttt{Filter}(C, S, \tau_s)$ \tcp{filter out if $s < \tau_s$}
	\tcc{extract appearance features}
	$\mathcal{F}_{det} \leftarrow \emptyset$ \;
	\ForEach{$\mathbf{x}$ in $C_{det}$}{
	$\mathbf{I}_{\mathbf{x}} \leftarrow \texttt{Crop}(f_k, \mathbf{x})$ \;
	$\mathcal{F}_{det} \leftarrow \mathcal{F}_{det} \cup H_{reid}(\mathbf{I}_{\mathbf{x}})$ \;
	}
	\tcc{hierarchical data association}
	Associate $\mathcal{T}$ and $C_{det}$ using distances of $\mathcal{F}_{trk}$ and $\mathcal{F}_{det}$ \;
	Associate remaining tracks and candidates using IoU \;
	$\mathcal{F}_{trk} \leftarrow \mathcal{F}_{trk} \cup  \mathcal{F}_{det}$ \;
	\tcc{initialize new tracks}
	$C_{remain} \leftarrow \text{remaining candidates from } C_{det}$ \;
	$\mathcal{F}_{remain} \leftarrow \text{features of } C_{remain}$ \;
	$\mathcal{T}, \mathcal{F}_{trk} \leftarrow \mathcal{T} \cup C_{remain}, \mathcal{F}_{trk} \cup \mathcal{F}_{remain}$ \;
}
	\caption{The proposed tracking algorithm.}
\end{algorithm}

\subsection{Analysis on Validation Set}
\label{sec:val}
\textbf{Contribution of each component.}
In order to demonstrate the effectiveness of the proposed method,
we investigate contribution of different components in our framework 
in Table \ref{table:val}.
The baseline method predicts new location of each track using the Kalman filter,
and then associates tracks with detection based on the IoU.
Using the classification probability to select candidates from both detection and tracks,
in which case, improves the MOTA by 4.6\%, comparing to the baseline method.
By punishing candidates from uncertain tracks,
the combination of tracklet confidence with classification probability
further improves the MOTA and reduces false positives,
as we expected in Section \ref{sec:conf}.
On the other hand, 
by introducing appearance representations based on ReID features,
we can obtain a significant improvement on the performance of identification (evaluated by IDF1 and IDS).
Our proposed method that combining the unified scoring function and ReID features 
has the best results for all metrics.

\begin{table}[t]
\small
\centering
\begin{threeparttable}
\caption{
Evaluation results on the validation set in terms of different components used.
\textbf{C:} classification probability, \textbf{T:} tracklet confidence, \textbf{A:} appearance representations with ReID feature.
The arrow after each
metric indicates that the higher ($\uparrow$) or lower ($\downarrow$) value is better.
}
\label{table:val}
\begin{tabular}{@{}l|lll|cccc@{}}
\toprule
\textbf{Method} & \textbf{C} & \textbf{T} & \textbf{A}  & \textbf{MOTA$\uparrow$} & \textbf{IDF1$\uparrow$} & \textbf{IDS$\downarrow$} & \textbf{FAF$\downarrow$} \\ \midrule
Baseline  &     &       &       &28.4     &32.8     &628      &0.85     \\
		&\checkmark &       &       &33.0     &37.6     &445      &0.77     \\
		&\checkmark &\checkmark &     &33.7     &37.3     &475      &0.63     \\ 
		&     &     &\checkmark &30.6     &42.4     &234      &1.01     \\ 
\midrule
Proposed  &\checkmark &\checkmark &\checkmark &\textbf{35.7}  &\textbf{45.3}  &\textbf{184} &\textbf{0.58}  \\ \bottomrule
\end{tabular}
\end{threeparttable}
\end{table}

\begin{table}[t]
\small
\centering
\begin{threeparttable}
\caption{
Evaluation results on the validation set in terms of different appearance representations.
}
\label{table:reid}
\begin{tabular}{@{}l|ccccc@{}}
\toprule
\textbf{Method} & \textbf{Length}  & \textbf{MOTA$\uparrow$} & \textbf{IDF1$\uparrow$} & \textbf{IDS$\downarrow$} & \textbf{FAF$\downarrow$} \\ \midrule
None      &-    &33.7     &37.3     &475      &0.63     \\
Color histogram &750  &34.9     &38.6     &250      &0.73     \\
HOG       &1152 &34.6     &38.5     &317      &0.70     \\
Color + HOG   &1902 &34.7     &39.3     &307      &0.68     \\
\midrule
ReID feature  &\textbf{512} &\textbf{35.7}  &\textbf{45.3}  &\textbf{184} &\textbf{0.58}  \\ \bottomrule
\end{tabular}
\end{threeparttable}
\end{table}

\begin{figure}[t]
\begin{center}
\includegraphics[width=0.95\linewidth]{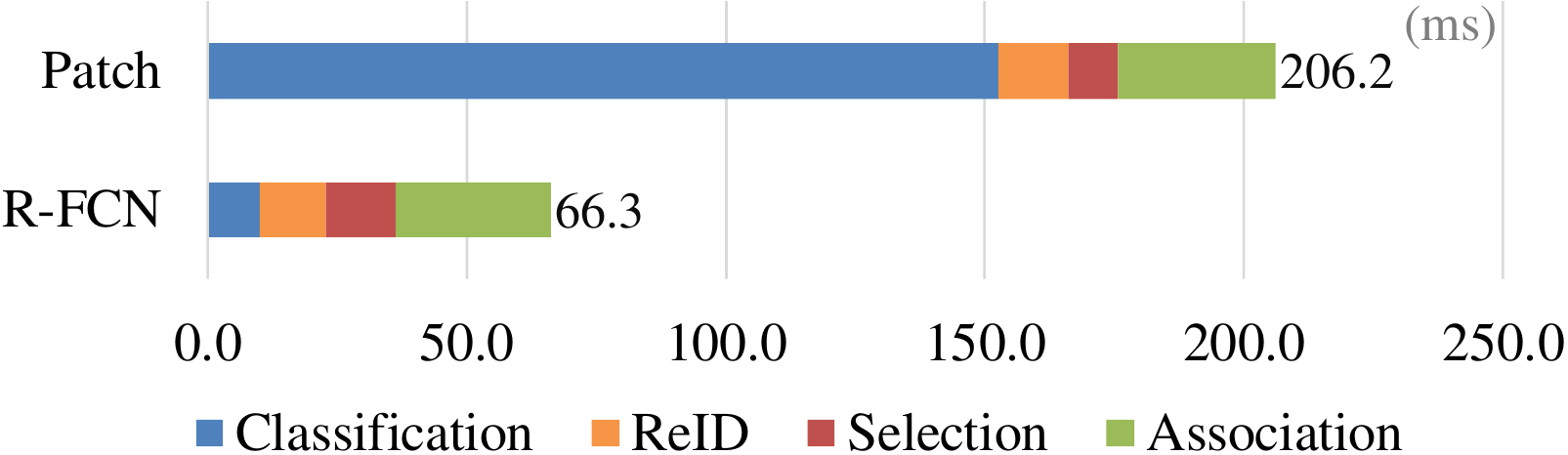}
\end{center}
\caption{
Average time consumption of one frame on MOT16-03 sequence, which contains over 50 people per frame.
\emph{Patch}: classification based on image patches using Squeezenet and two FC layers;
\emph{R-FCN}: classification using the same CNN backbone and a position-sensitive RoI pooling layer.
}
\label{fig:time}
\end{figure}

\textbf{Comparison with different appearance features.}
As shown in Table \ref{table:reid},
we compare representations learned by a data driven approach detailed in Section \ref{sec:reid}
with two typical hand-crafted features, 
i.e. color histogram, histogram of oriented gradient (HOG). 
Following the fixed part model, which is widely used for appearance descriptors \cite{satta2013appearance},
we divide each image of a person into six horizontal stripes with an equal size for the color histogram.
The color histogram of each stripe is built from the HSV color space with 125 bins.
We normalize both the color histogram and HOG features by $L^2$ norm,
and formulate the similarity using the cosine similarity function.
As shown in the table,
our appearance representation
outperforms traditional hand-crafted features by a large margin in terms of IDF1 and IDS,
in spite of the shorter feature vector comparing to other methods.
The evaluation result on the validation set verifies the effectiveness of 
our data driven approach for multiple people tracking.
The proposed tracking framework can be easily transfered for other categories,
by learning the appearance representation from corresponding datasets, 
such as vehicle re-identification \cite{liu2016large}.

\begin{table*}
\centering
\small
\begin{threeparttable}
\caption{Evaluation results on the MOT16 test set.}
\label{table:test}
\begin{tabular}{@{}lc|cccccccc|c@{}}
\toprule
\textbf{Tracker} & \textbf{Method} & \textbf{MOTA(\%)$\uparrow$} & \textbf{IDF1(\%)$\uparrow$} & \textbf{IDR(\%)$\uparrow$} & \textbf{MT(\%)$\uparrow$} & \textbf{ML(\%)$\downarrow$} & \textbf{FP$\downarrow$} & \textbf{FN$\downarrow$} & \textbf{IDS$\downarrow$} & \textbf{FPS$\uparrow$} \\ 
\midrule
LINF1 \cite{fagot2016improving} & batch & 41.0 & 45.7 & 34.2 & 11.6 & 51.3 & 7,896 & 99,224 & \textbf{430} & \textbf{4.2} \\
MHT\_DAM \cite{kim2015multiple} & batch & 45.8 & 46.1 & 35.3 & 16.2 & 43.2 & 6,412 & 91,758 & 590 & 0.8 \\
JMC \cite{tang2016multi} & batch & 46.3 & 46.3 & 35.6 & 15.5 & \textbf{39.7} & \textbf{6,373} & 90,914 & 657 & 0.8 \\
LMP \cite{tang2017multiple} & batch & \textbf{48.8} & \textbf{51.3} & \textbf{40.1} & \textbf{18.2} & 40.1 & 6,654 & \textbf{86,245} & 481 & 0.5 \\
\midrule
EAMTT \cite{sanchez2016online}  & \textbf{online} & 38.8 & 42.4 & 31.5 & 7.9 & 49.1 & 8,114 & 102,452 & 965 & 11.8 \\
CDA\_DDAL \cite{bae2017confidence} & \textbf{online} & 43.9 & 45.1 & 34.1 & 10.7 & 44.4 & 6,450 & 95,175 & 676 & 0.5 \\  
STAM \cite{chu2017online} & \textbf{online} & 46.0 & 50.0 & 38.5 & 14.6 & 43.6 & 6,895 & 91,117 & \textbf{473} & 0.2\\
AMIR \cite{sadeghian2017tracking} & \textbf{online} & 47.2 & 46.3 & 34.8 & 14.0 & 41.6 & \textbf{2,681} & 92,856 & 774 & 1.0 \\
\midrule
\textbf{MOTDT (Ours)} & \textbf{online} & \textbf{47.6} & \textbf{50.9} & \textbf{40.3} & \textbf{15.2} & \textbf{38.3} & 9,253 & \textbf{85,431} & 792 & \textbf{20.6} \\
\bottomrule
\end{tabular}
\end{threeparttable}
\end{table*}

\subsection{Evaluation on Test Set}
We first analyze the time consumption of the proposed tracking framework on MOT16-03 sequence.
As shown in Figure \ref{fig:time}, 
the proposed method is much more time efficient by sharing computations on the entire image. 

We report evaluation results on the test set of MOT16,
and compare our tracker with other offline and online trackers in Table \ref{table:test}.
Note that the tracking performance depends heavily on the quality of detection.
For the fair comparison, all the trackers in the table use the same detection provided by the benchmark.
As shown in the table,
Our tracker runs at real-time speed,
and outperforms existing online trackers on most of the metrics, 
especially for IDF1, IDR, MT, and ML.
The identification ability is enhanced by the deeply learned appearance representation.
The improvement on MT and ML demonstrates 
the advantage of our unified scoring function for candidate selection.
Selecting candidates from detection and tracks indeed reduces tracking failures caused by missing detection.
Moreover, our online tracker has much lower computational complexity
and is about 5\texttildelow 20 times faster than most of the existing methods.

\section{Conclusion}
In this paper, 
we propose an online multiple people tracking framework,
which takes full advantage of recent deep neural networks.
We tackle unreliable detection by selecting candidates from 
outputs of both detection and tracks.
The scoring function for candidate selection is formulated by an efficient R-FCN,
which shares computations on the entire image.
Moreover, we improve the identification ability when coping with intra-category occlusion
by introducing ReID features for data association.
ReID features trained by a data driven approach outperforms traditional hand-crafted features by a large margin.
The proposed tracker achieves real-time and state-of-the-art performance on the MOT16 benchmark.
A future study is planed to further improve efficiency by 
sharing convolutional layers with both classification and appearance extraction.

\begin{spacing}{0.98}

\bibliographystyle{IEEEbib}
\small
\bibliography{strings,refs}

\end{spacing}

\end{document}